\documentclass[acmsmall]{acmart}

\usepackage{silence}
\usepackage{caption}
\usepackage{fancyvrb}
\usepackage{multirow}

\usepackage[ruled]{algorithm2e} 

\usepackage{enumitem}

\usepackage{calc}
\usepackage{textgreek}

\usepackage{colortbl}
\usepackage{booktabs}
\usepackage{graphicx}
\usepackage[normalem]{ulem}
\useunder{\uline}{\ul}{}

\definecolor{plum}{rgb}{0.56, 0.27, 0.52}
\definecolor{darkred}{rgb}{0.55, 0.0, 0.0}

\setcopyright{acmlicensed}
\acmJournal{PACMHCI}
\acmYear{2021}
\acmVolume{5}
\acmNumber{CSCW2}
\acmArticle{350}
\acmMonth{10} \acmPrice{15.00}\acmDOI{10.1145/3476091}

\received{January 2021}
\received[revised]{April 2021}
\received[accepted]{May 2021}

\usepackage{mdframed}

\begin{document}

\title[Black or White but Never Neutral: How Readers Perceive Identity from Yellow or \newline \textbf{}Skin-toned Emoji]{Black or White but Never Neutral: How Readers Perceive Identity from Yellow or Skin-toned Emoji}

\author{Alexander Robertson}
\affiliation{%
  \institution{School of Informatics, University of Edinburgh}
  \city{Edinburgh}
  \country{Scotland}
  }
\email{alexander.robertson@ed.ac.uk}

\author{Walid Magdy}
\affiliation{%
  \institution{School of Informatics, University of Edinburgh}
  \city{Edinburgh}
  \country{Scotland}
  }
\email{wmagdy@inf.ed.ac.uk}

\author{Sharon Goldwater}
\affiliation{%
 \institution{School of Informatics, University of Edinburgh}
 \city{Edinburgh}
 \country{Scotland}
 }
\email{sgwater@inf.ed.ac.uk}

\renewcommand\shortauthors{Alexander Robertson, Walid Magdy and Sharon Goldwater}



\newcommand{\emoji}[1]{\includegraphics[height=1.40\fontcharht\font`\A]{images/emoji_images/#1.png}}

\newcommand{\changed}[1]{\textcolor{black}{#1}}

\begin{abstract}

Research in sociology and linguistics shows that people use language not only to express their own identity but to understand the identity of others. Recent work established a connection between expression of identity and emoji usage on social media, through use of emoji skin tone modifiers. Motivated by that finding, this work asks if, as with language, readers are sensitive to such acts of self-expression and use them to understand the identity of authors. In behavioral experiments (n=488), where text and emoji content of social media posts were carefully controlled before being presented to participants, we find in the affirmative---emoji are a salient signal of author identity. That signal is distinct from, and complementary to, the one encoded in language. Participant groups (based on self-identified ethnicity) showed no differences in how they perceive this signal, except in the case of the default yellow emoji. While both groups associate this with a White identity, the effect was stronger in White participants. Our finding that emoji can index social variables will have experimental applications for researchers but also implications for designers: supposedly ``neutral`` defaults may be more representative of some users than others.

\end{abstract}

%
%


%
%

\maketitle

\section{Introduction}\label{section:intro}

Emoji, such as \emoji{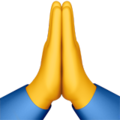} \emoji{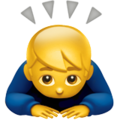} \emoji{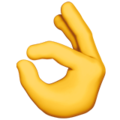}, are icons used to enrich computer-mediated communication. First available in carrier-specific versions on Japanese mobile phones in the 1990s, they became a formally recognized part of the Unicode Standard in 2010 with v6.0 including 847 individual icons. By 2015, almost 40\% of Instagram posts contained an emoji \cite{dimson_2015} and of half a billion Twitter posts in 2017, 14\% contained an emoji \cite{robertson2018}.

In 2015, recognizing that ``people all over the world want to have emoji that reflect more human diversity'' \cite{tsr51}, version 8.0 of the Unicode Standard added five new codepoints -- modifiers which could only be used in conjunction with specific emoji. These modifiers change the appearance of the target emoji, rendering it with a more human-like skin tone rather than the standard bright yellow: \emoji{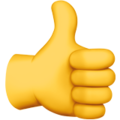} could now also be \emoji{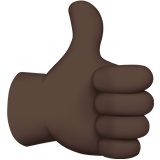} \emoji{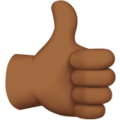} \emoji{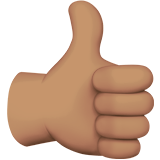} \emoji{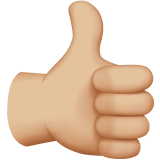} \emoji{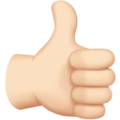}. The five modifiers, based on the Fitzpatrick Phototyping Scale \cite{fitzpatrick}, cover a range of skin tones from darker to lighter. 

At the time, commentators feared that these new emoji would be used in negative or racist ways \cite{bbcblackface,techcrunchblackface}, if they were used at all \cite{guardiandiverse}. However, these fears proved unfounded. On the contrary, \citet{robertson2018} showed that skin-tone modifiers are used widely, with no evidence of abusive use. A later study \cite{robertson2020} also found that in Twitter users across four diverse sample groups\footnote{\citet{robertson2020} sampled four groups of English-speaking Twitter users with human profile photos: from London, New York, Johannesburg, and a random sample. The three geographic locations were chosen to be culturally and demographically distinct.}, a user's most commonly used emoji skin-tone modifier closely matched the skin tone shown in their profile photo, and users with darker skin tones were more likely to have used skin-tone modifiers. \citet{robertson2018,robertson2020} conclude that tone-modified emoji are used as a form of self-representation.

Put another way, these findings suggest that  skin-tone modifiers are part of how users construct their online identity. Identity construction (whether online or not) is a process that occurs within a particular sociocultural context and defines who we are in relation to others: ``the human capacity -- rooted in language -- to know `who's who' (and hence `what's what')'' \cite[p.6]{jenkins2014social}. Online communication provides different affordances and challenges for identity construction compared to traditional communication \cite{boyd2007youth}, but in both traditional \cite{labov1972sociolinguistic,campbell2010naturesociopercep} and online \cite{bamman2014gender,dong2016survey,shoemark2017aye} settings, people actively construct their identity (or identities) by making choices---whether self-consciously or not---about how they present themselves \cite{labov1972sociolinguistic,campbell2009nature,bucholtz2005identity}.

Since identity is only meaningful as a relationship with others, it is important to understand not only how people \emph{produce} language and other artefacts to construct identity, but also how others \emph{perceive and interpret} those acts. 
Yet there has been little work so far in this area in terms of online communication. This paper begins to fill that gap, inspired by work on spoken communication from the field of sociolinguistics. That work has shown that linguistic cues (ranging from detailed acoustic/phonetic characteristics to lexical and syntactic usage) can carry social meaning---i.e., that listeners use such cues to infer aspects of the speaker's identity, such as their regional background, class or sexual orientation  \cite{labov1972sociolinguistic,smyth2003male-sexuality}. Listeners also use linguistic cues to judge how  intelligent \cite{campbell2009nature}, credible \cite{levari-credibility}, prestigious \cite{beinhoff-identity-accent} or trustworthy \cite{kinzler-accent-trust} a speaker seems. These cues alone are salient enough to provoke reactions to the speaker's perceived identity, such that the outcome can be negative and even harmful. Prior work has shown how this can result in loss of opportunity \cite{baugh1996perceptions-housing,hopper1973speech-employability}.

We hypothesize that, just as with spoken language, individuals use emoji both to express their own identity and to understand the identity of others. Furthermore, as has been demonstrated for spoken language, we hypothesize that this understanding can affect how readers react to content containing emoji -- for example, how likely they are to believe it or to pass it on. These potential behavioral consequences have clear practical implications in areas such as marketing, politics, and the spread of disinformation. However, we focus here on the necessary foundation for any follow-up studies on these behavioral aspects and ask:

\begin{itemize}
    \item [$\bullet$] Do readers indeed infer aspects of an author's identity (specifically, ethnicity) via their emoji usage (specifically, use or non-use of emoji skin-tone modifiers)?
    \item [$\bullet$] If so, do these inferences depend on the reader's own identity, and how does the emoji signal interact with other (linguistic) cues to identity?
\end{itemize}

\changed{Both the nature of social categories (here, ethnicity) and the potential indicators of those categories (here, emoji skin-tone modifiers and Tweet text) depend strongly on cultural context. Since ours is the first work of its kind, we did not attempt a fully general study. Instead, we limited our participant group to native English-speakers living in London, to ensure a relatively common cultural and linguistic context, and we focused our experiments on just two identity categories, Black and White (both for participants' self-identified ethnicity, and the categories they were asked to infer).
This study design neglects participants who identify as neither Black or White, nor can it directly inform us about how emoji might be perceived to identify such individuals. However, for this initial study of emoji perception, we decided to use the two categories that we felt would be (to these participants) most typically associated with particular skin tones---Black with the darker end of the Fitzpatrick scale, and White with the lighter end. Our hypothesis (borne out by our experiments) is that the skin tones of emoji, like their real-life counterparts, are associated with particular identities, and are used by readers to infer the author's identity. It will be important to follow up our study by investigating how emoji skin tones are used and/or perceived by individuals who identify with other social/ethnic  groups.}

\changed{Based on the discussion above, in} the remainder of this paper, we use the terms ``Black'' and ``White'' to refer to the (self-identified) ethnicities of our participants and the categories to which they assign imagined authors.\footnote{Although there are popular dictionary-derived delineations between the terms ``race'' and ``ethnicity'' (e.g. race is physical, ethnicity is cultural \cite{natgeoraceeth}), these obscure the socially-constructed nature of both, as considered within the framework of Critical Race Theory \citet{crt-hci}. We use the term ethnicity here for consistency with the self-identity question on the participant recruitment platform, which referred to ethnicity.} \changed{We follow \citet{blodgett2016demographic} in referring to particular signals as ``demographically-aligned'', using the terms ``Black-aligned'' and ``White-aligned''} to refer to texts and emoji that are associated with these two groups. The alignment of texts is determined through a norming study (described below), and the alignment of emoji is based on the Fitzpatrick Scale labels assigned by the Unicode Consortium, with the two darkest tones considered ``Black-aligned'' and the two lightest ones ``White-aligned''. We do not use the middle tone.

Our main findings, \changed{based on a set of controlled behavioral experiments (n=488) which were pre-registered with the Open Science Framework\footnote{Available online at https://osf.io/tn8mg/}} are as follows:

\begin{description}[leftmargin=!,labelwidth=\widthof{\textbf{3}}, labelindent=1cm, rightmargin=1cm, itemsep=0.2cm]
\item [1] Skin-toned emoji provide a salient signal of author identity and readers readily perceive this. 
\item [2] This signal is complementary to any signal encoded in the other linguistic content of a social media text, meaning that the appropriate emoji can either boost an existing signal (if the signals are congruent) or dampen it (if they are not).
\item [3] We find no evidence of differences in how Black and White readers perceive Black-aligned and White-aligned emoji, despite a known asymmetry in how Black and White authors (in general, and in London specifically) produce these emoji on social media \cite{robertson2020}.
\item [4] The yellow emoji is not perceived as neutral by either Black or White readers. On average, both groups perceive it as more likely to index a White identity, and we find this effect to be stronger among White readers.
\end{description}

The first two findings confirm that in terms of their social meaning, emoji act in many ways like linguistic communication. We hope this will inspire further work aiming to generalize our findings to groups from other cultural or linguistic backgrounds, draw out subtler effects, and explore the consequences of readers' inferences on their social media behavior. 

Our finding that yellow emoji are not viewed as neutral is also important for designers considering how to implement systems that offer equal opportunities for user representation. This finding supports recent claims (discussed in more detail below) that even when a supposedly neutral technological option is provided, it may unwittingly build in (or take on) assumptions that create an unbalanced system which better suits one group of users over others. Our study also shows how claims of neutrality can, at least in some cases, be tested experimentally, and we hope that this will inspire designers to consider similar approaches at earlier stages of the design process.

\section{Background and Related Work}\label{section:background}

As indicated above, our work draws on, and has implications for, research in multiple areas. In this section, we briefly review three of these and discuss their relation to our own work. We begin by discussing research on identity and its expression on social media, which our study expands upon to look specifically at emoji. Next, we review key work in sociolinguistics, focusing in particular on the experimental methodology developed there for exploring how people perceive identity from communication, and which we use in our own study. Finally, we review work on design and bias in technology, which bears especially on our findings regarding the supposedly neutral yellow emoji.

\subsection{Self-representation, Social Media and Emoji}

Identity and self-representation are important aspects of human behavior, \changed{with the work of \citet{goffman1959presentation} being most influential in our understanding of self-representation as a human act. Under Goffman's theatrical metaphor, people are actors who perform different roles in society for different audiences, using props and costumes to do so. These can be literal props and costumes, such as the tools and clothing we use when playing the role associated with our employment. In the context of social groups, they may take the form of owning specific items or dressing in a particular style. But performing a particular role can also involve our behavior, such as facial expressions, language and voice.} 

As new technologies emerge, novel methods of social networking develop, \changed{providing new audiences and roles to play, new ways to play existing roles, and new (perhaps metaphorical) props and costumes. For example,} \citet{facetedidentity2011} examined how American internet users express facets of their identity (e.g. work and family) and found that different technologies (such as email and social media) were preferred for different social purposes, based on factors such as privacy and intimacy. \citet{hogan2010presentation} argued that social media users ``perform'' the self through ``exhibitions'': carefully curated usage of website functionality such as profiles, avatars, friends and followers. Again, \changed{the user's relationship with their imagined audience}
was an important consideration for people, especially given the fact that a single platform may bring together multiple audiences --- the notion of ``context collapse'' \cite{boyd2007youth}.

Various researchers have studied how an individual's social group or identity can affect how they use website and social media functionality. These studies have mainly focused on individuals' use of sites or production of content, through quantitative analysis of their usage in relation to some demographic factor such as gender or age. \citet{papacharissi2012without} examined the content associated with hashtags on Twitter, which highlighted the role of linguistic creativity and flexibility in expressing the self online. \citet{kapidzic2015race} looked at profile photos, showing gender and race differences in how teenagers presented themselves in profile photos (e.g. posture, clothing). \citet{JORDANCONDE2014356} looked at late-stage adolescents and how they experiment with their identity on social media, while \citet{wood2016digital} argue that social media is an important playground for adolescents of all ages who are, perhaps self-consciously, still developing their identity. People can express multiple aspects of a changing identity through social media \cite{hogan2010presentation} during transitional periods other than age, for example moving away to college \cite{collegetransition} or gender transition \cite{haimsonsocialtransitionmachinery}. In such instances social media acts as ``social transition machinery'', affording users the opportunity to experiment with and reconstruct their identity. While all of these works highlight the importance of social media in constructing identity, they have primarily focused on the producer, rather than the perceiver, as we do here.

Turning to emoji and identity, quantitative work has shown stark differences not only in emoji usage by men and women \cite{chen2018genderlens} but also in the relative prevalence of the five emoji skin tones \cite{barbieri2018gender,coats2018skin}. By manually annotating Twitter accounts for skin tone based on profile photos, \citet{robertson2020} found that users, regardless of geographical location, generally employ skin-toned emoji which match their profile photo skin tone, strongly suggesting a self-representational function in production. The use of particular emoji in Twitter bios has been shown to closely track personal traits and tastes, which are shared across groups of users \cite{li-etal-2020-emoji}. Meanwhile, studies of how emoji are {\em interpreted} have focused more on their semantics and pragmatics than their social signaling function, but have nevertheless found differences in interpretation across social groups. For example, \citet{herring2020agegender} looked at interpretation based on the age and gender of the reader, and found an age differential, with younger people more able to understand conventionalized pragmatic functions of emoji. More generally, the semantics of emoji have been inventoried as part of the EmojiNet project \cite{wijeratne2016emojinet,wijeratne2017emojinet}, while \citet{novak2015sentiment} used human ratings of Tweets to determine the sentiment of different emoji. The fact that emoji are rendered differently depending on which platform they are viewed \cite{miller2016,miller2017,miller2018} has been shown to cause confusion for readers when it comes to interpreting their pragmatic or semantic functions. In these studies, we note that the focus is generally learning what the reader thinks about the emoji. The work to be presented here takes a different view, focusing on what a reader thinks about an author --- the emoji is not explicitly brought to the attention of the reader.

Identity and representation are intricately linked with culture. This has been explored widely in sociology and cultural studies. The work of Hall \cite{jackson2010encyclopedia} argued that the generation of meaning through language is mediated by complex cultural processes. Such meaning shapes human identity and develops within a cyclical, self-reinforcing system yet it is not fixed, despite any perceived immutability or centrality to our existence. Looking at emoji as cultural artefacts, \citet{freedman2018cultural} claims that emoji are wholly dependent on cultural interpretations in order to be viewed as discriminatory or representational: in a vacuum, emoji are ``unreadable''. Studying emoji usage from different cultural perspectives reveals both similarities and differences between East (China, Japan) and West (Canada, United Kingdom, United States) countries \cite{guntuku2019studying}. Emoji usage in general was similar in all areas, as were the distributional semantics of emoji. Differences were mainly observed in the emoji that occurred in contexts relating to food, leisure and friendship.

\subsection{Experimental Sociolinguistics}

Linguistic variation has been studied at the geographic level, as in work on dialectal differences between regions of England \cite{hughes2013english} or varietal differences in languages shared between countries such as England and New Zealand \cite{gordon2004new}. Of more relevance to our research here is foundational work in sociolinguistics, showing how these variations can be used to index social variables \cite{labov1972sociolinguistic}. These linguistic cues can form a key component of identity, marking someone as a member or non-member of a particular group or class of people. These works establish the fact that people \emph{produce} language to express their own identity. More recently there has been a focus on how our \emph{perception} of a speaker's identity is influenced by the language of other people. As outlined in Section~\ref{section:intro}, we hypothesize that identity can \emph{also} be perceived through emoji.

To test this we use a variant of the Matched Guise Test \cite{lambert1960evaluational}. This experimental paradigm is commonly used in sociolinguistics to explore attitudes of groups towards a specific aspect of language. Participants are assigned to groups based on some known variable (e.g. gender, age, level of education) and exposed to stimuli which have been strictly controlled in order to exhibit a particular linguistic feature (e.g. particular vowel sounds, specific words, syntactic constructions). All other variation (e.g. speed, volume, pitch) is kept constant. The participant, using a questionnaire, states their opinion of the person represented by the stimulus -- age, where are they from, social status, intelligence, or some other aspect of research interest. The aim is to determine whether \changed{participants, on average,} are sensitive to linguistic signals and if they connect these with non-linguistic properties. 

The majority of works in this area employ auditory stimuli, but text stimuli can be used to elicit perceptions for suitable variables. For example, \citet{staum2010listeners} investigated perceptions of word-final t/d deletion, since this variable can be directly expressed in writing. Other studies have used text-based experiments to examine participants' perceptions based on lexical diversity \cite{bradac1988lexical} or the use of quotative ``like'' versus ``go'' \cite{buchstaller2005putting}. In all cases, the use of text stimuli made it easier to control linguistic variation and remove possible confounding factors in the stimuli.  Our exact experimental procedure is described in more detail in Sections~\ref{section:norming} and~\ref{section:exp_setup}.


\subsection{Technology Design and Bias in Emoji}

There is considerable evidence that technology, rather than being neutral, actually incorporates systematic biases that favor some groups (often, white men) over others (e.g. women \cite{perez2019invisible} and non-white people \cite{algopp2018,barbosa2019rehumanized,raji2019actionable}). One way in which bias surfaces is in the representation of identity, where White is overwhelmingly the default or only option \cite{brockgames2011,robotsrace2020,cave2020whiteness}.
When technology is explicitly non-White, as in the case of Blackbird, a web browser aimed at serving the needs of Black internet users, reactions are highly critical---such technology is often judged to be unnecessary, because existing solutions are considered to be ``neutral''\cite{brock2011beyond}.

The introduction of emoji skin-tone modifiers was met with similar reactions. The yellow emoji available in 2015 was considered by some to be neutral and representative of anyone \cite{viceyellow,wapoyellow}, so there were claims nobody would use the skin-toned versions \cite{guardiandiverse}. These negative reactions did not emerge only upon the release of the skin-tone modifiers. As shown by \citet{miltner2020one}, in a succinct summary of debate on the Unicode Consortium's internal mailing lists, the addition of skin tones to the Unicode Standard was not a smooth process. Even the order in which the skin tones should be encoded (from lightest to darkest, or from darkest to lightest?) was a point of contention --- some spoke out against the ``over-correctness'' of a dark-to-light ordering. Miltner concludes that ``the racial composition of the original emoji set was ultimately shaped by an institutionalized form of colorblind racism which insists that concerns regarding racial representation and identity are irrelevant to ``neutral'' technical systems''. 

Beyond the technical specification of emoji, vendors must implement their physical design. Most appear to have taken a ``white first'' \cite{whitefirstdesign} approach by creating the White versions and then adjusting the palette, rather than designing each version individually. This is best illustrated by the addition of the foot emoji (\emoji{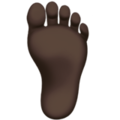} \emoji{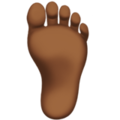} \emoji{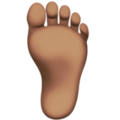} \emoji{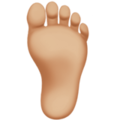} \emoji{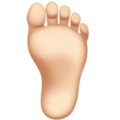} \emoji{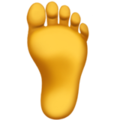} ) in 2018 which confused Twitter users \cite{buzzfeedfeet} with its dermatological inaccuracy --- the soles of feet (and the palms of hands) are less melanistic than the rest of a person's skin, but this is not readily obvious to those with generally less melanistic skin. Even before 2015, some vendors had \emph{only} white-skinned emoji --- see the Emojipedia entry for Apple's iOS6 emoji\footnote{https://emojipedia.org/apple/ios-6.0/} for one example. 

Despite such issues in design and media speculation regarding their usefulness, \citet{robertson2020} found that users on Twitter with darker skin, regardless of their geographic location, were more likely to have used skin-toned emoji than users with lighter skin. They also used such emoji more often than other users. This suggests that yellow is not neutral. One aim of our work here is to provide firm evidence as to whether yellow emoji are indeed neutral or not.

\section{Hypotheses and Research Questions}\label{section:hrq}

This paper tests three main hypotheses and several additional research questions inspired by the work discussed above. As noted in Section~\ref{section:intro}, the hypotheses and research questions (as well as details of the experiment and analysis) were pre-registered with the Open Science Foundation.\footnote{Available online at https://osf.io/tn8mg/} Our main hypotheses deal with how skin-toned or yellow emoji are interpreted by readers, when the emoji occurs with a \emph{non-aligned} text: i.e., a text that is neither Black-aligned or White-aligned according to our norming study.

\begin{description}[leftmargin=!,labelwidth=\widthof{\textbf{Hypothesis 3}}, labelindent=0cm, rightmargin=0cm, itemsep=0.2cm]
\item[Hypothesis 1] Readers of a non-aligned text containing a Black-aligned or White-aligned emoji will more likely attribute authorship to a Black or White author.
\end{description}

This hypothesis tests whether readers use emoji as a salient cue to author identity. It is motivated by long-standing results from sociolinguistic studies showing that different social variables are indexed by specific linguistic variables \cite{labov1972sociolinguistic} and that people are aware of these associations both consciously and subconsciously \cite{d2018controlled}.

\begin{description}[leftmargin=!,labelwidth=\widthof{\textbf{Hypothesis 3}}, labelindent=0cm, rightmargin=0cm, itemsep=0.2cm]
\item [Hypothesis 2] Black readers will be more likely than White readers to attribute authorship to someone whose skin tone is similar to that of the emoji, given a non-aligned text containing a Black-aligned or White-aligned emoji.
\end{description}

The second hypothesis is motivated by the finding that people with darker skin tones produce more skin-toned emoji in their social media posts \citep{robertson2018}. We predict that this greater experience of expressing identity through emoji will result in Black participants in our experiments being more attuned to perceiving such emoji use, by others, as an explicit act of self-representation.

\begin{description}[leftmargin=!,labelwidth=\widthof{\textbf{Hypothesis 3}}, labelindent=0cm, rightmargin=0cm, itemsep=0.2cm]
\item [Hypothesis 3] Black readers will be more likely to attribute the default yellow emoji to a White than a Black identity, given a non-aligned text.
\end{description}

Claims that the yellow emoji is neutral and represents everyone run counter to evidence from Twitter data \cite{robertson2020} that Black users, regardless of geographic location, are less likely to use the yellow emoji than White users. When presented with a text containing a yellow emoji, we predict Black users will more often judge the author to be White than Black. White users, who use the yellow emoji more often, should therefore see the yellow emoji as representing neither a Black nor a White identity.

In addition to our main hypotheses we have more general research questions. These compare the relative impacts of the identity signal encoded in text and emoji. Compared to a baseline of a Black-aligned or White-aligned text with no emoji, does adding an emoji result in a ``boosting effect'', wherein readers more often perceive the author as having an identity matching the combined text/emoji signal? Or is there no effect, possibly because one signal is already overwhelmingly salient? How do readers respond to conflict between the text and emoji signals, such as a Black-aligned text with a White-aligned emoji? Do they resolve the conflict by choosing one signal over the other, or do the signals moderate each other?
We might expect some differences between groups, since the two groups may not be equally familiar with the linguistic characteristics of Black-aligned or White-aligned text, and Black readers may be more familiar with self-representational usage of emoji because of their popularity among Black Twitter users \cite{robertson2018}. Therefore, Black users may be more willing to ignore the text signal in favor of the emoji signal.

\begin{description}[leftmargin=!,labelwidth=\widthof{\textbf{Research Question 3}}, labelindent=1cm, rightmargin=1cm, itemsep=0.2cm]

\item [Research Question 1] What effect is observed when the identity signals encoded in text and emoji are congruent? 

\item [Research Question 2] What effect is observed when the identity signals encoded in text and emoji are incongruent, creating a conflict between the two? 

\item [Research Question 3] How do group differences in linguistic/emoji knowledge affect how signal conflict is resolved?

\end{description}

\section{Norming study}\label{section:norming}

\changed{As noted in the introduction, both the delineation of social categories and the interpretation of cues to those categories depend heavily on cultural context. Therefore, any study connecting the two must be careful to choose participants who are likely to share a common cultural context, and thus a shared interpretation of the relevant social categories and how they might be communicated. To address this issue, our main study recruited participants from an online platform (Prolific.co) whose profile listed them as native English-speakers living in London, with ethnicity listed as either Black or White.\footnote{The platform asks "What ethnic group do you belong to?" with possible responses being Asian, Black, Mixed, White, or Other.)} We also use the terms Black and White as labels for the identities that we ask participants to infer in our experiments, based either on Tweet text alone, or text with emoji. }

In order to present participants in the main study with tweets that would have real characteristics of Black or White authors from the same cultural context as the participants, we selected the stimuli using a norming study.\footnote{We are keenly aware that the terms Black and White will not have a universally agreed-upon interpretation, since these are socially constructed categories \cite{crt-hci,delgado2017critical}. Their precise meaning to different individuals, and the extent to which individuals identify with these labels, may vary even within a specific region such as London. Similarly, even within London, many different dialects of English are spoken, and linguistic factors can signify not only ethnic identity but also age, class, and other social variables. Nevertheless, we assume that our participants will have some degree of shared understanding both of the categories Black and White, and of the linguistic features that tend to associate with those categories. Although there may be some differences between participants, our grouped design abstracts away from these differences. Investigating individual differences in the effects we find here is an important avenue for future work, but beyond the scope of our study.}
Using the Twitter 1\% random sample API in 2017 ("the 2017 dataset"), we first gathered tweets from users based in London (according to their profile location text). We then identified the skin tone of the authors by crowd-sourced annotation of their Twitter profile photos. Each photo was labeled by three annotators, by making a comparison to the five-point emoji skin tone range. We kept only those authors where all three annotators agreed on the label. From authors with darker or lighter skin we selected 50 tweets each, re-sampling until tweets contained no profanity, sensitive topics or personal information, while preserving the ratio of dark-skinned to light-skinned authors.\footnote{While not all of these authors would necessarily present or identify as Black or White, we rely on the norming study (below) to select tweets from this set that do index these identities.} All tweets ended with a skin-toneable emoji (e.g. \emoji{thumb5}, \emoji{thumb4}, \emoji{thumb3}, \emoji{thumb2}, \emoji{thumb1} or \emoji{thumb}) and contained no other emoji.

To determine the extent to which the text alone is informative about author identity, we use a modified version of the Matched Guise Test introduced in Section~\ref{section:background}, following \citet{staum2010listeners}. The reader is shown a tweet with the emoji removed. Rather than complete a multi-point questionnaire as in the standard Matched Guise Test \cite{lambert1960evaluational}, the reader is forced to make a simple binary choice --- is the author of the text Black or White? There is no option for the reader to say they are unsure or to skip a stimulus. When unable to decide, they must pick at random. The outcome is the ratio of Black:White judgments for each stimuli. This ratio will be approximately 50:50 if participants as a group do not make any association between the linguistic properties of the text and author identity. However, if there is such an association, this will skew the ratio of Black:White judgments in the appropriate direction.

\changed{The use of a forced binary choice, with no possibility to express uncertainty, is widely used in research into human decision making processes and is better known in experimental psychology as the two-alternative forced-choice task. In that context, the method is often used to understand decision making processes, with a dependent variable such as accuracy or reaction time 
\cite{ratcliff1998modeling,bogacz2006physics}. 
However, the use of a binary choice with no option to express neutrality or uncertainty also 
makes it possible to detect small effects when results are aggregated over participants. This is especially important for effects that may be subconscious, such as attitudes towards, and preconceptions of, language or identity (as in both the norming and main study, where we use the same forced choice methodology).}

Ethics approval was obtained from the University. Each consenting participant was shown all 100 texts, \changed{in randomized order}. 40 participants in total were recruited through the Prolific platform. Prolific allows researchers to recruit participants based on a wide range of voluntarily-declared demographics that they provide in their profile. We recruited 20 participants who self-identified as Black and 20 who identified as White. The outcome was 40 ratings per text, from which the ratio of Black:White judgments for each was calculated. Participants were paid £2.25 (hourly rate: £8.20).

\begin{figure}
    \centering
    \includegraphics[width=\textwidth]{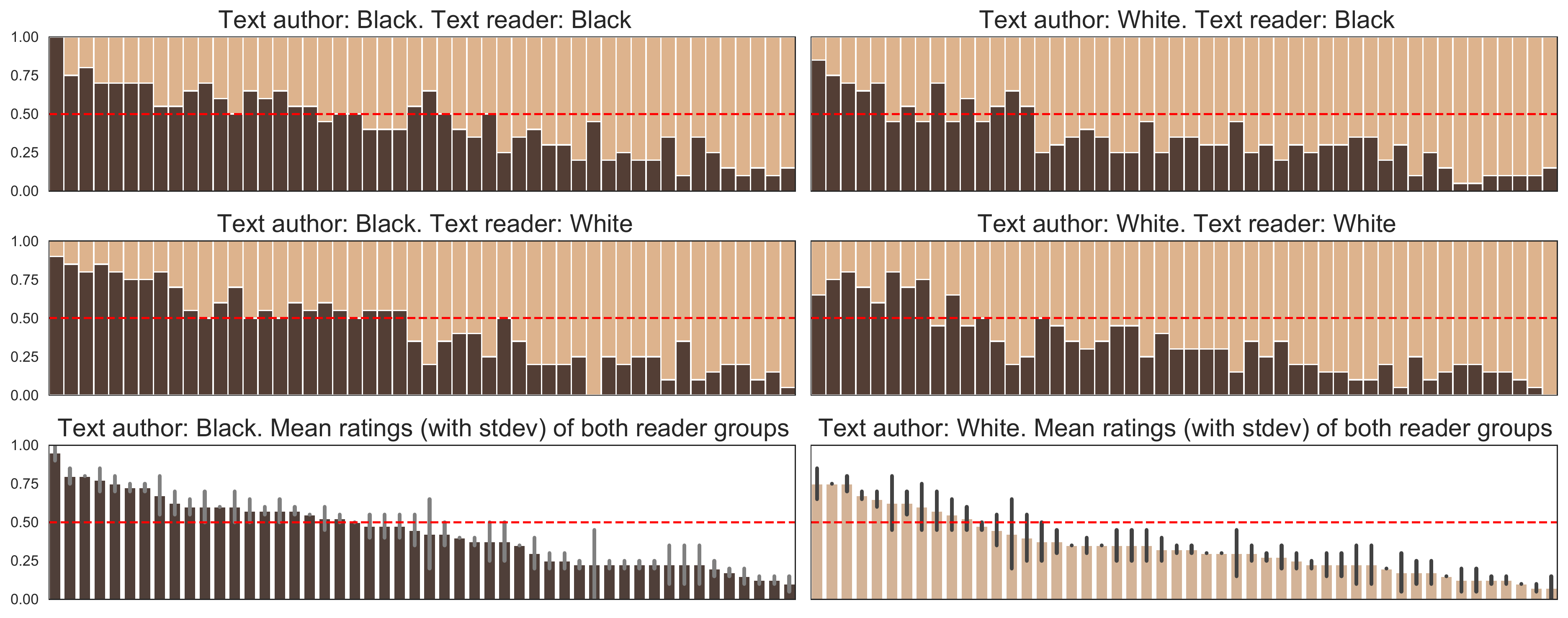}
    \caption{Results of the text norming study. Each bar is a text, comparing the ratio of Black (dark brown) to White (light brown) judgments. Rows 1 and 2 show the ratio of Black:White judgments for all four combinations of author/reader ethnicity. Row 3 shows the mean ratio of each text, combining Black and White reader judgments. Texts are ordered by mean ratio. Standard deviation bars denote the extent to which the two reader groups agree in their judgments for a given text. Red line denotes the point of at-chance judgment.}
    \label{figure:norm_results}
\end{figure}

In Figure~\ref{figure:norm_results}, we show the results for each of the 100 texts. Many are judged similarly by both Black and White participants, but the standard deviation for the proportion of White judgments for some texts is as high as 0.32. Mean standard deviation for all stimuli is 0.09 (\textsigma=0.07).

\begin{figure}
    \centering
    \includegraphics[width=\textwidth,trim=125 0 125 0]{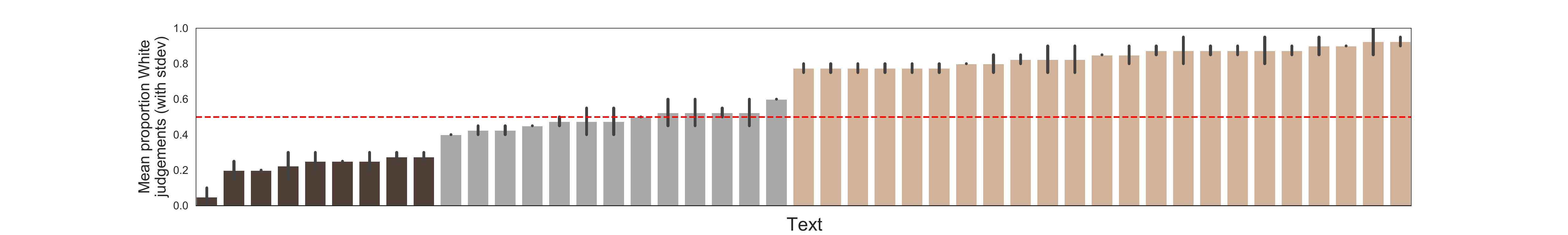}
    \caption{Mean proportion of White authorship judgments of final stimuli meeting selection criteria, with standard deviations showing variation between judgments of Black and White readers. Red line denotes the point of at-chance judgment. Bar color denotes classification of stimuli based on proportion of White judgments by all readers: Black-aligned text is $\leq$ 0.3, non-aligned text is between 0.4 and 0.6, White-aligned text is $\geq$ 0.7.}
    \label{figure:final_texts}
\end{figure}

To classify texts as being Black-aligned, non-aligned or White-aligned we used a simple heuristic based on the level of agreement between Black and White participants. Using the aggregate ratio over all participants would introduce a confound in our experiments --- we want each stimuli to be perceived the same way by all participants. Therefore, Black-aligned/White-aligned texts are those where over 70\% of \emph{both} Black and White participants rated the text as having a Black/White author. Non-aligned texts are those rated as having Black authorship by between 40\% and 60\% of \emph{both} Black and White Participants. \changed{45 tweets did not meet any of these criteria and were discarded. This left 9 Black-aligned texts, 13 non-aligned texts and 23 White-aligned texts.} Figure~\ref{figure:final_texts} shows the mean ratio of judgments per text by all participants. The mean standard deviation for these texts is 0.05 (\textsigma=0.04), with a maximum of 0.11. This subset of stimuli therefore exhibit a higher degree of agreement between Black and White participants than the whole set.

\section{Experimental setup}\label{section:exp_setup}

From the norming study, we randomly selected 8 Black-aligned, non-aligned and White-aligned texts, for a total of 24 texts. These are shown in the Appendix, with a single example laid out in Table~\ref{tab:stim_generation}. Each text was manipulated to create 4 critical stimuli differing only in terms of their emoji content for a total of 96 stimuli --- one with no emoji, one with a Black-aligned emoji, one with a White-aligned emoji, one with a yellow emoji. These 96 stimuli were arranged into 4 blocks of 24. Each block contained each sentence only once, with each of the four emoji manipulations appearing six times. Participants were assigned to a block at random. These stimuli allow us to measure a response variable (whether a participant selects Black or White) in relation to one of three explanatory variables: reader ethnicity (Black or White), linguistic bias of the text (Black-aligned, non-aligned, White-aligned), or the emoji content of the text (none, Black-aligned, yellow or White-aligned).

\begin{table}[h]
\centering
\resizebox{\textwidth}{!}{%
\begin{tabular}{@{}lll@{}}
\toprule
\multicolumn{1}{c}{Use} & \multicolumn{1}{c}{Details} & \multicolumn{1}{c}{Example} \\ \midrule
Source data & The original Tweet & Next step manager then owner. By 28 just watch me \emoji{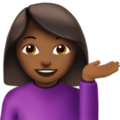} \\
Norming study & Used to determine linguistic bias & Next step manager then owner. By 28 just watch me \\
Main study & Critical stimuli, no emoji & Next step manager then owner. By 28 just watch me \\
Main study & Critical stimuli, Black-aligned emoji & Next step manager then owner. By 28 just watch me \emoji{tip4} \\
Main study & Critical stimuli, Yellow emoji & Next step manager then owner. By 28 just watch me \emoji{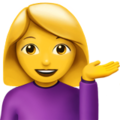} \\
Main study & Critical stimuli, White-aligned emoji & Next step manager then owner. By 28 just watch me \emoji{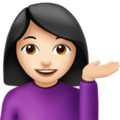} \\ \bottomrule
\end{tabular}%
}
\caption{Example of stimuli generated from a single Tweet, written by a Black author, ending with a tone-modifiable emoji. The text alone was considered linguistically non-aligned (approximately equal numbers of Black and White judgments) by participants of the norming study.}
\label{tab:stim_generation}
\end{table}

In order to obfuscate the focus of the experiment (risking participants simply matching their response to the emoji), 48 filler trials were included. The filler trials asked participants to choose between a female/male or an old/young author identity, rather than a Black/White identity. The text was provided by random tweets (containing no profanity, sensitive topics or personal information) from the 2017 dataset. The four emoji types were also balanced across the fillers to prevent the critical trials being conspicuous due to emoji only appearing when the choice was between Black or White. Each participant therefore completed 72 trials (48 filler, 24 critical) \changed{--- 6 critical trials with no emoji, 6 with Black-aligned emoji, 6 with White-aligned emoji and 6 with yellow emoji}. These stimuli were presented to participants using the same modified Match Guise Test \cite{staum2010listeners} as the norming study, \changed{again in randomized order}. The only difference is that participants were presented with a binary response question about the age, sex or ethnicity of the author (depending on whether the trial was critical or filler), and stimuli could also contain emoji.\changed{As with the norming study, there was no option for participants to say they are unsure or to skip a trial and they are instructed to select at random when they are unable to decide. The outcome is the ratio of Black:White judgments for each stimuli. As per \citet{staum2010listeners}, the ratio will be approximately 50:50 if participants \emph{as a group} do not make any association between the linguistic properties of the text and author identity. However, if there is such an association, this will skew the ratio of Black:White judgments in the appropriate direction.}

As in the norming study, we obtained ethics approval from the University. We recruited 488 participants via the Prolific.co platform, based on their self-reported demographic details. None had taken part in the norming study. All were native English speakers and based in London, to maximize familiarity with the linguistic content of our stimuli. Half the participants self-identified as Black, half as White, as determined by demographic information provided by Prolific. This number of participants was selected based on a power analysis conducted with the G*Power software package \cite{gpower}. This sample size gives our statistical analysis (using Fisher's Exact Test) a power of 0.9 to detect an effect size of 0.15 (that is, the difference of proportions between two measures being tested) with the standard 0.05 alpha error probability. Being able to detect a smaller effect size or achieve greater power would require between 500 and 700 participants in each group. This was not feasible, given the low numbers of Black participants registered on the Prolific.co platform. We therefore aimed for the greatest power/smallest effect size possible. For statistical analysis, to determine whether there is a significant difference in the the ratios of judgments for stimuli based on factors such as emoji type, text type or participant ethnicity, we use Fisher's Exact Test and report odds ratios and $p$-values.

Participants who indicated they may not have been paying due attention by completing the experiment too slowly or too quickly (±2 standard deviations from the mean completion time of 9.9 minutes (\textsigma=8.8) for Black participants, 6.7 minutes (\textsigma=2.3) for White), or who simply pressed the same response key on every trial, were replaced with new participants prior to any analysis being performed. This amounted to 14 participants excluded for taking too long: 6 Black and 8 White. Excluded participants were still paid the standard rate for their time: £1.56 for a ten minute experiment (hourly rate: £9.36).

\section{Results and Analysis}\label{section:results}

In all cases, we generate a 2x2 contingency table for the response variable (whether the reader selected Black or White) with respect to one relevant explanatory variable (reader ethnicity, text type, emoji type), with the other two held constant. For example, to analyze the effect of Black-aligned and White-aligned text on responses, we would generate a 2x2 table for each reader ethnicity group. Rows represent Black and White judgments, the columns represent the two text alignments. Therefore, cells report the number of each judgment type for each of the text alignments. Analyzing this table using Fisher's Exact Test determines if there is a statistically significant difference in the proportions of the values of the response variable, relative to each value of the explanatory variable.

\subsection{Baselines}

\begin{figure}
    \centering
    \includegraphics[width=\textwidth]{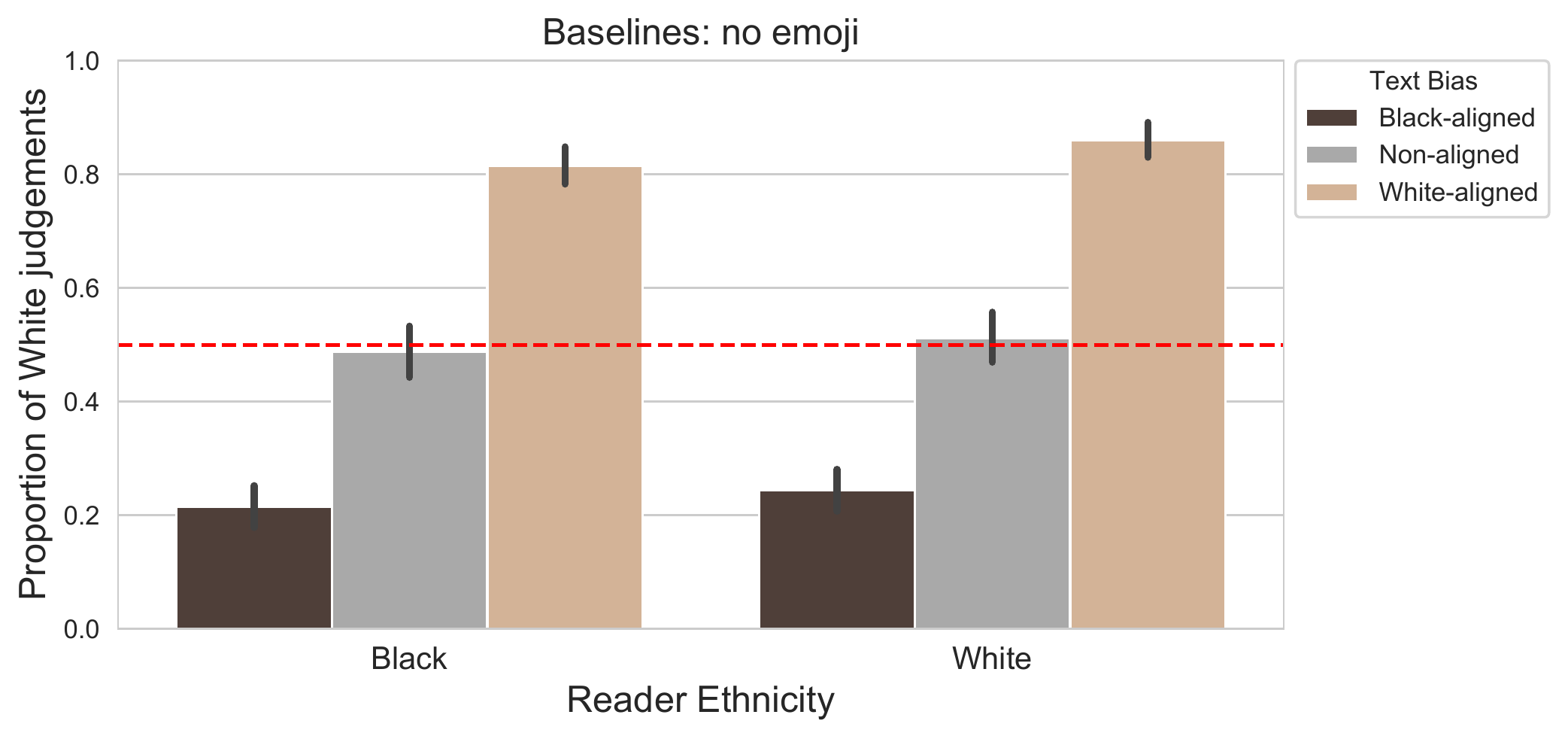}
    \caption{Baseline results: Proportion of White judgments (with 95\% CI) for different text types, with no emoji added. Red bar denotes judgments at chance.}
    \label{figure:baseline_results}
\end{figure}

We first report baseline results in Figure~\ref{figure:baseline_results}. These are reader judgments of Black-aligned/non-aligned/White-aligned texts containing no emoji. The non-aligned texts are judged at chance by both participant groups (Black: 48.77\%; White: 51.23\%), while the ratio of judgments for Black-aligned texts (Black: 21.52\%; White: 24.39\%) and White-aligned texts (Black: 81.56\%; White: 86.07\%) is skewed in favor of the text type.

\subsection{Hypotheses 1 and 2: How readers interpret skin-toned emoji}

\begin{figure}
    \centering
    \includegraphics[width=\textwidth]{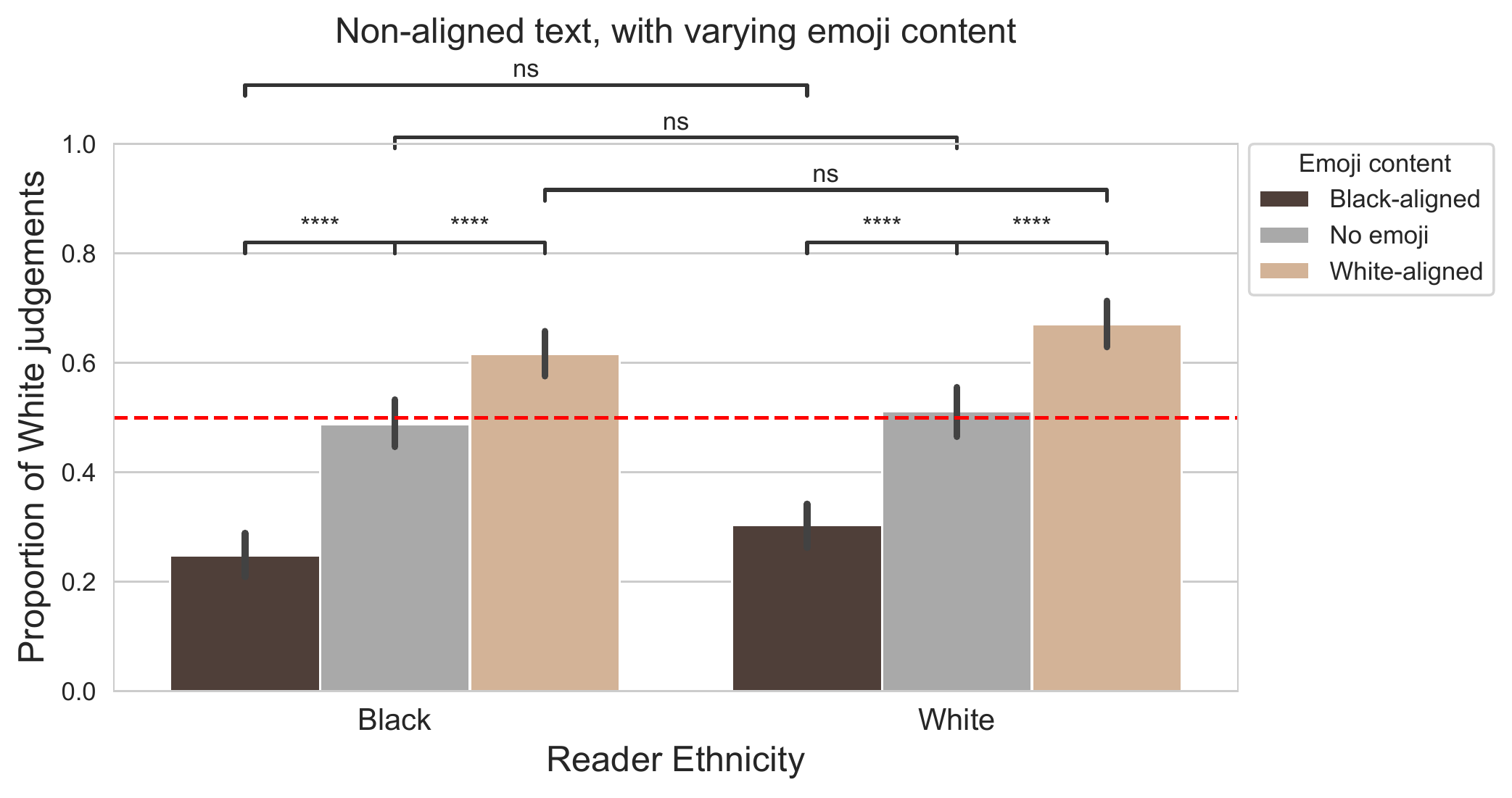}
    \caption{Proportion of White judgments (with 95\% CI) for non-aligned texts with Black-aligned or White-aligned emoji. Non-aligned texts with no emoji are shown, as baseline reference. H1 compares within participant groups; H2 between participant groups. Red bar denotes judgments at chance. Brackets denote the interval that $p$-value falls within: ns: (0.05, 1]; *: (0.01, 0.05]; **: (0.001, 0.01]; ***: (0.0001, 0.001]; ****: $p$ $\leq$ 0.0001.}
    \label{fig:h1_h2_results}
\end{figure}

Hypothesis 1 tests whether readers perceive skin-toned emoji as salient indicators of author identity, while Hypothesis 2 tests whether this perception is different in Black and White reader groups. Figure~\ref{fig:h1_h2_results} shows results for both hypotheses. The ratio of judgments by Black readers for non-aligned texts containing emoji is significantly different from their baseline (Fisher's exact test. Black-aligned emoji: odds ratio=2.887, $p$=9.0693e-15; White-aligned emoji: odds ratio=1.691, $p$=6.4329e-05). White readers are also significantly different from their baseline (Fisher's exact test. Black-aligned emoji: odds ratio=2.413 $p$=3.9385e-11; White-aligned emoji: odds ratio=1.934, $p$=7.0562e-07). Comparing Black and White reader groups reveals no significant differences for Black-aligned nor White-aligned emoji (Fisher's exact test. Black-aligned emoji: odds ratio=1.32, $p$=0.062412; White-aligned emoji: odds ratio=1.262, $p$=0.094715).

Accordingly, the data support Hypothesis 1 but not Hypothesis 2 --- readers do use Black-aligned and White-aligned emoji to determine authorship, but no evidence was found to support a claim that Black and White reader groups use emoji in this way to different extents. This finding may seem at odds with the results of Robertson et al.'s \cite{robertson2020} emoji production study, which found that Twitter users with darker skin tones are both more likely to use skin-toned emoji in the first place and to use them more often. A possible explanation for the discrepancy is differences in the importance placed on self-representation by Black and White users of social media.\footnote{Other explanations are possible, for example that White users are more likely than Black users to use emoji to refer to other people rather than to themselves. There is evidence that users are reluctant to use skin-toned emoji when representing others \cite{robertson2020}, so this could explain White users' lower rate of skin-toned emoji. However the differences between White and Black users' rate of self-referential emoji would need to be rather large, so this explanation seems unlikely.}
Such differences were also observed between ethno-racial groups in a study of students on Facebook: in comparison with White or Vietnamese users of Facebook, students of African American, Latinx or Indian heritage projected ``a visual self that is dramatically more social'' \cite{ethnofacebook2009}. A key component of that study was focus groups and in-depth interviews and such methods will be necessary for any future work which aims to provide a more complete view of why, rather than how, people express their identity through emoji.

A further analysis (not included in the pre-registration) suggests that although we found no evidence of difference between Black and White reader groups, there is a difference in how reader groups perceived the Black-aligned and White-aligned emoji. Specifically, when aggregating both reader groups (since there is little difference between them), we found that adding a Black-aligned emoji to neutral text pushes the proportion of Black judgments from approximately 50\% to, on average, 72.44\% (\textsigma=3.91). For a White-aligned emoji, White judgments make up 64.34\% (\textsigma=3.77) of responses. The deviation from random chance is not symmetric for both emoji: on average, a Black-aligned emoji elicits a stronger response of ~23\%, compared to ~14\% for a White-aligned emoji.

One account for this disparity is that Black-aligned emoji are more strongly associated with authentic expression of identity, exactly because they are produced for that reason \cite{robertson2018,robertson2020}. An alternative account is that Black-aligned emoji, which we presented on a light background, are just visually more prominent in our experiments. However, we doubt this alternative account based on a follow-up study where we measured reaction times to emoji. 58 participants were recruited through social media, receiving gratitude as payment. They were shown 37 text stimuli from our experiments - 19 fillers with no emoji, 6 neutral stimuli each with a Black-aligned, White-aligned or yellow emoji. Participants pressed a key as quickly as possible if there was an emoji in the text. We discarded trials over 5000ms. Overall accuracy was 90.5\%. Reaction times for correct Black-aligned trials (\textmu=810ms, \textsigma=475) and White-aligned trials (\textmu=826ms, \textsigma=532) were not significantly different (Independent samples t-test. t=-0.39, $p$=0.70).

\subsection{Hypothesis 3: How readers interpret yellow emoji}

\begin{figure}
    \centering
    \includegraphics[width=\textwidth]{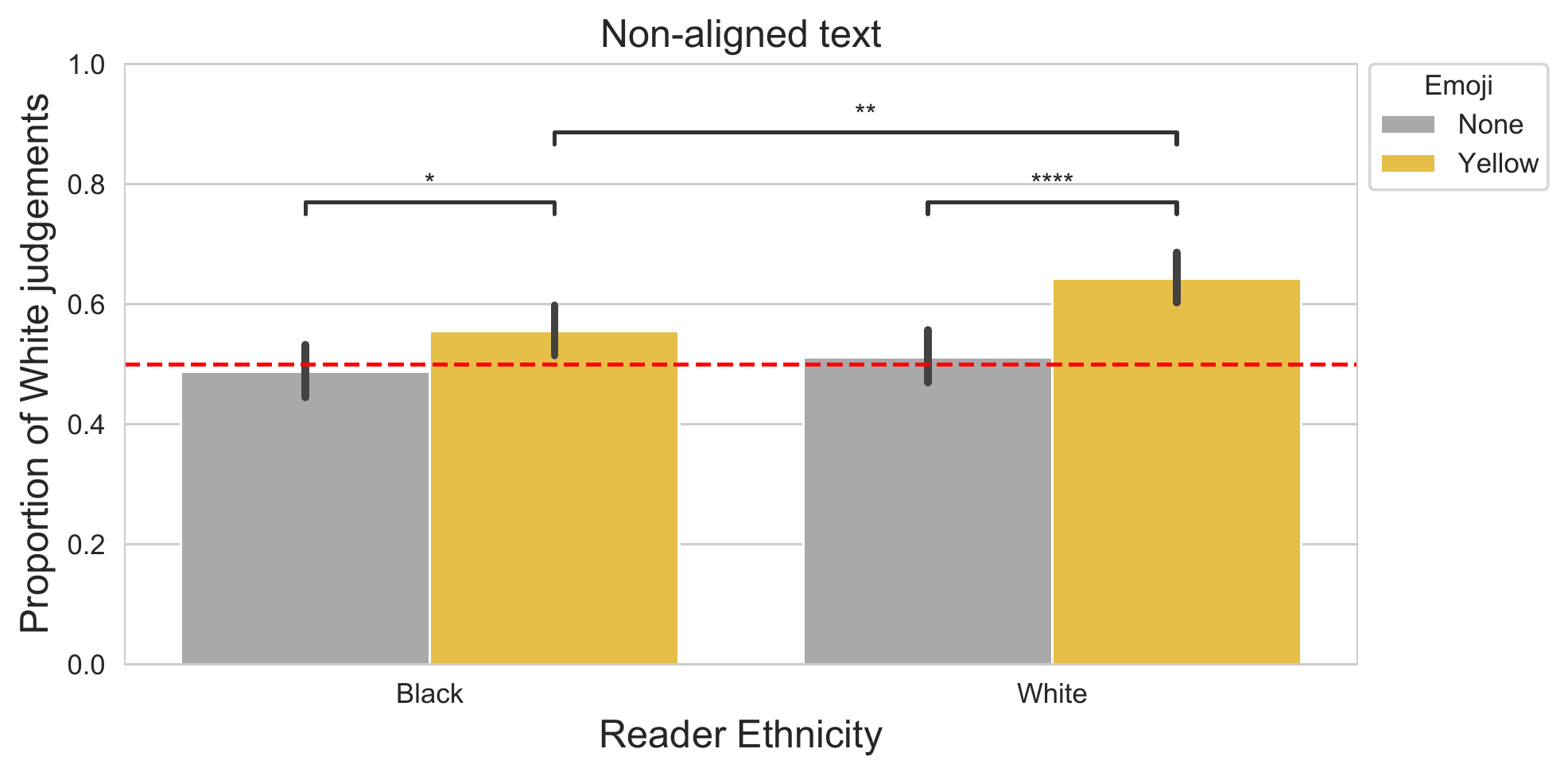}
    \caption{Proportion of White judgments (with 95\% CI) for non-aligned texts with yellow or no emoji. Red bar denotes judgments at chance. Brackets denote the interval that $p$-value falls within: ns: (0.05, 1]; *: (0.01, 0.05]; **: (0.001, 0.01]; ***: (0.0001, 0.001]; ****: $p$ $\leq$ 0.0001.}
    \label{fig:h3_results}
\end{figure}

Hypothesis 3 tests whether Black readers perceive the yellow emoji as more representing a White than a Black identity, with results shown in Figure~\ref{fig:h3_results}. This proved to be the case, with non-aligned texts with yellow emoji being perceived significantly differently from the baseline of non-aligned text with no emoji (Fisher's exact test. Black readers: odds ratio=1.31, $p$=0.04). We performed the same test for White readers and found that they also saw such texts as representing a White identity (Fisher's exact test. White readers: odds ratio=1.72, $p$=4.38e-05). This runs counter to our prediction that only the Black reader group would see yellow as White.

Although RQ3 (discussed later) focused on how Black and White reader groups compare when viewing Black-aligned or White-aligned emoji, we performed a similar analysis here for the yellow emoji\footnote{This additional analysis was not explicitly included in the pre-registration document.}. There is a significant difference in how Black and White reader groups perceive authorship of neutral text containing a yellow emoji (Fisher's exact test: odds ratio=1.445, $p$=0.006). Both rate the yellow emoji as representing a White author, with this effect more pronounced in White readers (+13.11\% over baseline) than in Black (+6.76\% over baseline).

These results confirm that Black readers see the allegedly neutral yellow emoji as representing a White identity and that this perception is also held by White readers, and to a greater extent. This difference is not due to a wide margin between baselines: Figure~\ref{figure:baseline_results} shows that these are essentially equivalent across the two reader groups. Instead, we suggest that readers may be more likely to perceive yellow as White simply because the yellow and lighter emoji tones are visually more similar than the yellow and darker tones\footnote{The visual similarity can be quantified by representing the colors within the CIELAB model \cite{schanda2007colorimetry} and using the Delta E CIE 2000 algorithm \cite{doi:10.1002/col.20070} to measure their difference. CIELAB was specifically designed such that changes in a color's values (which measure how black/white, green/red and blue/yellow a color is) directly correspond to the visual perception of those changes by humans. The larger a Delta E value between two colors, the larger the perceptual difference. Delta E is 58.36 for Black/yellow and 19.38 for White/yellow.}. Although emoji have only recently become more human-like, it seems that even their cartoonish origins may have (however inadvertently) created a situation of ``white first'' design \cite{whitefirstdesign}, as discussed in Section~\ref{section:background}.

\subsection{Research Questions 1 and 2: The interaction of emoji and language signals}

\begin{figure}
    \centering
    \includegraphics[width=\textwidth]{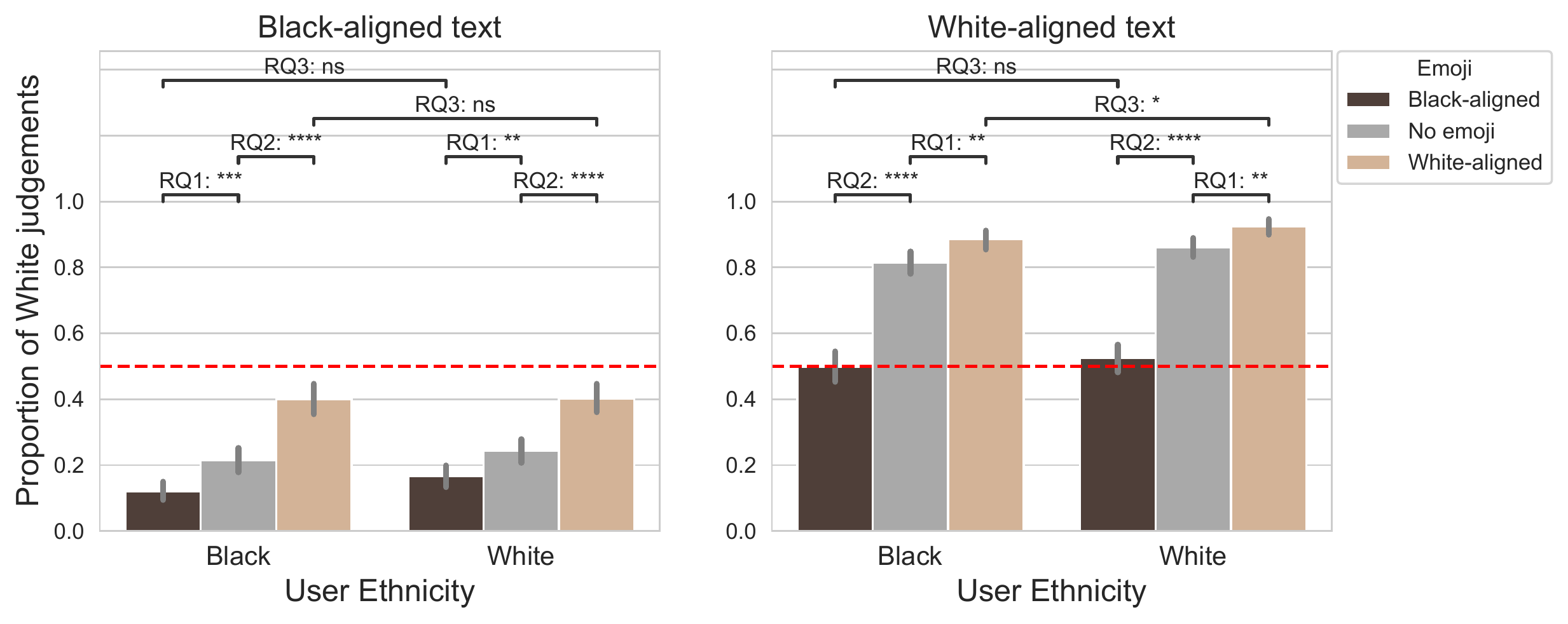}
    \caption{Proportion of White judgments (with 95\% CI) for Black-aligned text (left) or White-aligned text (right), depending on reader ethnicity and whether a Black-aligned, White-aligned, or no emoji is present. RQ1 and RQ2 test within participants, for congruent and incongruent text/emoji combinations, respectively. RQ3 tests between participant groups for both congruent and incongruent. Red bar denotes judgments at chance. Brackets denote the interval that $p$-value falls within: ns: (0.05, 1]; *: (0.01, 0.05]; **: (0.001, 0.01]; ***: (0.0001, 0.001]; ****: $p$ $\leq$ 0.0001.}
    \label{fig:rq_results}
\end{figure}

Research questions 1 and 2 examine how different text/emoji combinations affect judgment of author identity. Adding a similarly-aligned emoji to a text (RQ1: Figure~\ref{fig:rq_results}) results in a boosting effect: compared to a baseline of the text with no emoji, there is a significantly different proportion of judgments in favor of the alignment. For each of the congruent combinations, change over baseline was significant for both Black readers (Fisher's exact test. Black-aligned/Black-aligned: odds ratio=1.99, $p$=0.0001; White-aligned/White-aligned: odds ratio=1.74, $p$=0.00297) and White readers (Fisher's exact test. Black-aligned/Black-aligned: odds ratio=1.62, $p$=0.0032; White-aligned/White-aligned: odds ratio=1.97, $p$=0.0018). The effect of emoji alone is slightly lower than that of text alone (as per Figure~\ref{figure:baseline_results} and Figure~\ref{fig:h1_h2_results}), but we find a ``boosting effect'' when both signals are present and in agreement. From this we conclude that each signal is distinct and neither one overpowers the other.

When text/emoji alignment are mismatched (RQ2: Figure~\ref{fig:rq_results}), there is instead a damping effect relative to the text-only baseline: 
for both reader groups, mismatched text/emoji results in judgements that are significantly different from the text-only baseline in the direction that is closer to random
(Black-aligned text, White-aligned emoji - Fisher's exact test.
Black readers: odds ratio=2.43, $p$=5.3505e-10;
White readers: odds ratio=2.08, $p$=1.7704e-07.
White-aligned text, Black-aligned emoji - Fisher's exact test.
Black readers: odds ratio=4.46, $p$=6.3499e-26;
White readers: odds ratio=5.60, $p$=8.2273e-31)). 

For the particular texts we used, the White-aligned text/ Black-aligned emoji condition shows both reader groups at chance for judging author identity \changed{(the first and fourth bars in the right-hand figure)}, while the opposite arrangement is slightly below chance \changed{for both groups (the third and sixth bars in the left-hand figure)}. This suggests the two conflicting signals do not simply cancel each other out, and it is likely that the final resolution (how strongly the reader perceives a particular identity) depends on factors such as what specific linguistic cues are present and how strong they are---factors which we did not carefully control for in our study. As per Section~\ref{section:hrq}, we selected texts based only on their overall classification by participant groups in the norming study. \changed{
Therefore, it could be either that Black-aligned texts in general are a more salient indicator of author identity than White-aligned texts, or simply that the particular texts we used had this property.} Careful manipulation of the linguistic content from which text alignment is derived could give a better understanding of what generated this result. 

How readers behave prior to making their judgment could also be informative about difficulty of resolving text/emoji incongruity. In psycholinguistics, eye-tracking of participants faced with ambiguity is highly informative as to what extent the reader is aware of conflicts \cite{rabagliati2017children}. Reaction times are commonly used as a proxy for cognitive load: responding to ambiguous words in sentences takes longer \cite{simpson1994context}. We tentatively report that mean reaction times were reduced when text/emoji were White-aligned/Black-aligned, compared to Black-aligned/White-aligned. Black readers were 2.6s slower (8.7s vs 6.1s), White readers 0.3s (5.0s vs 4.7s). We stress, however, that the experimental paradigm here is not suited to measuring reaction times: we did not control for stimuli length or restrict stimuli presentation time. We offer this analysis only to encourage potential future work on the increased cognitive load that incongruent emoji/text may, or may not, provoke.

\subsection{Research Question 3: How reader groups compare in perceiving congruent and incongruent signals}

Results are again shown in Figure~\ref{fig:rq_results}, marked with RQ3. Comparing results between Black and White readers shows a significant difference in how they judge texts in the White-aligned/White-aligned congruent setting (Fisher's exact test. White-aligned Text/White-aligned Emoji: odds ratio=1.16, $p$=0.049) but not the Black-aligned/Black-aligned setting (Fisher's exact test. Black-aligned Text/Black-aligned Emoji: odds ratio=1.45, $p$=0.054). Neither of the incongruent settings show a statistical difference between groups (Fisher's exact test. Black-aligned Text/White-aligned Emoji: odds ratio=1.11, $p$=0.44; White-aligned Text/Black-aligned Emoji: odds ratio=1.01, $p$=1.0).

This result is only just past the threshold of statistical significance, with White readers in this particular condition having a slightly higher proportion of congruent judgments than Black readers (0.924 vs 0.885). Similarly, the result for two Black-aligned signals only just misses the threshold of statistical significance ($p$=0.055). It is possible that our sample size is too small to reliably detect any small effect that may or may not occur here. If there is a very small but significant difference between Black and White readers in their own congruent conditions, it may be because each group is especially sensitive to a matching text/emoji signal which also matches their own identity, which is triggered by in-group preferences being especially well-catered to in these conditions. However, calculating the Bayes factors for contingency tables \cite{bayesfactor} to compare Black and White readers in each congruent condition (Bayes Factor: Black-aligned congruent=0.42; White-aligned congruent=0.40) suggests there is anecdotal evidence for \emph{no association} between the explanatory variable (reader ethnicity) and the response variable (their judgment of the author's identity).


\section{Discussion}\label{section:discussion}

Overall, we found that, when our participants read an otherwise neutral text, the presence of a skin-toneable emoji was enough to sway them (on average) toward ascribing a particular ethnic identity to the author: Black if the emoji had a dark skin tone, and White if it had either a light skin tone or no tone at all (yellow). Moreover, when the text also contained socially meaningful information about the author's ethnic identity, the effect of this information was moderated by the emoji, indicating that text and emoji provide complementary sources of information.

\changed{
Our study was necessarily limited in scope, with findings derived from a single set of participants within a particular region. Moreover, } our use of the labels Black and White (as well as the grouped design of the study) glosses over many possible differences between participants in terms of how they understand these terms with respect to themselves and others. 
\changed{Therefore, even within London, some individuals may not demonstrate the same effects as the group average we found, and we don't know the extent to which groups from other regions would show the same effects for the toned emoji (especially if the Black/White distinction is less socially meaningful, or differently delineated, in that region). Nevertheless, we have shown that both social media text and emoji \emph{can} be used to infer aspects of identity, and that in our experiment, they \emph{were}. Given the many related sociolinguistic studies indicating that social signaling through language is probably universal, we would argue that even based on just this one study, it is likely that
social media users who are experienced with emoji are inferring aspects of identity from them all the time. We would also expect that, as we found here, emoji and text provide complementary signals.}
Furthermore, we would expect that for regions of the world where Black and White \emph{are} relevant categories associated with skin color, qualitatively similar results would be found regarding their association with dark vs light emoji tones; and that in other regions, the emoji skin tones may signal other socially relevant categories. We hope that our work will inspire follow-ups to confirm these predictions, as well as to investigate potential individual differences. 

Although we did not find any between-group differences in the perception of ethnic identity from emoji, such differences might still be found across other groups or for other aspects of identity. Existing work has shown that readers of different ages and genders differ in their interpretations of the pragmatic \citep{herring2020agegender} and emotional \citep{agediffemojieval2020} functions of emoji---i.e., the author's intentions. It seems likely that there can also be differences in the interpretation of social meaning, especially since on social media, authors and readers may share much less social context than speakers and listeners usually do. Therefore, an important future step would be to consider what happens when the social signal in an emoji is perceived in ways that differ from its intended meaning.

Our experiments also addressed the question of whether the default yellow emoji is in fact ``neutral''. Among our participant groups, we found that it was not. 
We suggested one possible reason is that the yellow color is more visually similar to the lighter tones associated with the White identity than to the darker tones associated with the Black identity. However, it is also possible that the association with White has less to do with visual similarity than with the fact that the yellow emoji is the default. Within the British socio-cultural context, where White is the historically dominant and ``default'' category and other ethnicities are seen in contrast to it (see work on the experiences of migrants to the UK and what it means to be White in Britain \cite{limitsofwhiteness,fox2015denying,halvorsrud2019maintenance}), the yellow emoji may associate with this identity precisely because of its default status. In that case, we might expect that in other parts of the world with different demographics or different social hierarchies, the yellow emoji would be associated with other groups. Preliminary evidence from Robertson et al.'s \cite{robertson2020} study of emoji production suggests that demographics alone may not be enough to change the association between yellow and White: they found that even in Johannesburg, where the majority of sampled user profile photos have dark skin, those users were more likely to have used non-yellow emoji tones than lighter-skinned users. This suggests that yellow does not simply represent a statistical default, but that its color and/or other socially motivated assumptions lead to a stronger association with lighter-skinned and/or White people than with darker-skinned and/or Black people. Additional experiments with participants from a range of regional and cultural backgrounds could provide further evidence on the matter.

A related issue is whether the perceived Whiteness of the yellow emoji is a \emph{consequence} of the introduction of the other skin tones, or whether it already had that connotation beforehand. This question is difficult to answer in retrospect, but since our paper highlights a clear issue that could arise with other self-representations (a supposedly neutral representation that either isn't, or transitions away from neutrality following other changes), we hope that designers or researchers might follow up by studying some future change both before and after it occurs, to discover whether and how user perceptions change as a result of new options. While it is clearly important for users to feel they have options for self-representation, our results suggest that it may be difficult to provide such options without inadvertently advantaging one group over another. Nevertheless, considering these issues in advance can at least reduce the chances of perpetuating ``white first'' design \cite{whitefirstdesign}. We recommend designers consider our empirical findings within the context of Critical Race Theory for Human Computer Interaction \citep{crt-hci}.

In addition to these implications for research and design, our findings also have broader social implications. The inferences that a reader makes about an author's identity may change how the reader feels or acts toward the author, or the author's content. This can have both positive and negative ramifications. A recent example is found in a study by \citet{fitzpatrick}, of players in a trust-based investment game who were permitted to communicate either through plain text or text with emoji. Emoji users saw greater financial gains from partners, compared to non-users. And although the author set out to explore the impact of emoji usage in general, the results showed evidence that usage of gendered/skin-toned emoji particularly impacted trust in participants. In a more sinister vein, there is recent evidence \cite{freelon2020black} of ``troll farms'' engaging in a form of ``digital blackface'' \cite{bbcblackface} in order to further their agenda by spreading disinformation through accounts made to look like Black activists. \citet{freelon2020black} measured digital blackface through manual assessment of screen names, profile descriptions and tweet content. Given \changed{our finding} that people use emoji to index identity and social categories, an open question now is whether this can be used to manipulate readers in the same way that profile photos and screen names have been.

Finally, there are implications for end users. We found that using skin-toned emoji on social media provides a strong signal from which one's ethnicity (if not other social categories) may be inferred --- especially if it matches other linguistic signals. Twitter users may already be sensitive to this, as seen by the phenomenon of tone modulation whereby users deliberately turn skin tones off for specific messages \cite{robertson2020}. Although that analysis considered users who already displayed a profile photo of themselves, the same study also showed that users without a profile photo used skin-toned emoji to the same extent as those with a public photo on display. For all of these users, simply determining the most common emoji skin tone used in their online messages would reveal their ethnicity --- a possibility which may alarm users who deliberately avoid using a profile photo. 
Our work demonstrates yet another example of how social media data may provide information about users that they have not declared \changed{directly} --- for example, a user's stance on particular topics can be inferred from network analysis \cite{abeerstance}, public tweets aimed at private accounts can result in leakage of potentially personal information \cite{10.1145/3394231.3397919}, and a user's views on certain political issues are associated with particular linguistic choices, even when not discussing that issue \cite{shoemark2017aye}. Users would quite rightly be disturbed if Twitter had a ``search by skin color'' function, but emoji arguably facilitate the determination of exactly this information. One option is for platforms to disallow search for skin-toned emoji or to include \emph{all} toned variants in search results, to prevent identification and targeting of specific groups of users.

To sum up, our study presented (as far as we know) the first experimental evidence that emoji skin tones are used by readers to infer aspects of the author's identity. This finding has wide-ranging implications for the study of sociolinguistics, online identity, and the spread of disinformation, but also raises many further questions which we have outlined above. Our work adopted the Matched Guise methodology from experimental sociolinguistics, with a pre-registered experimental design. We hope that by demonstrating these methodologies, our work encourages other computational social scientists to use them too, whether to follow up on this study or to pursue other questions. Finally, our results regarding yellow emoji highlight the fact that even supposedly neutral options carry social meaning, which may advantage some groups over others. It is not clear how to avoid this problem entirely, but we hope that by raising awareness, our work can inform the development of new self-representational technologies in the future.

\section{Acknowledgments}

This work was supported in part by the EPSRC Centre for Doctoral Training in Data Science, funded by the UK Engineering and Physical Sciences Research Council (grant EP/L016427/1) and the University of Edinburgh.

\bibliography{main_2021_cleaned}
\bibliographystyle{plainnat}

\newpage
\appendix
\section{Appendix}\label{section:appendix}

\begin{table}[h]
\centering
\resizebox{\textwidth}{!}{%
\begin{tabular}{@{}lrcccc@{}}
\toprule
ID & Text & Bias & Black & Yellow & White \\ \midrule
\rowcolor[HTML]{EFEFEF} 
1 & Big tune after Big tune & Black & \emoji{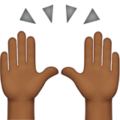} & \emoji{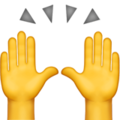} & \emoji{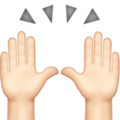} \\
2 & Finally got that fresh cut & Black & \emoji{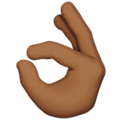} & \emoji{ok} & \emoji{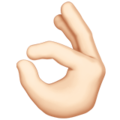} \\
\rowcolor[HTML]{EFEFEF} 
3 & I dont sip the syrup, i got friends to lean on & Black & \emoji{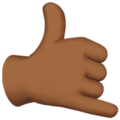} & \emoji{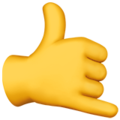} & \emoji{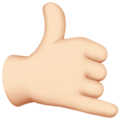} \\
4 & If you don't feel like your girl a 10 then idk & Black & \emoji{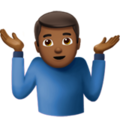} & \emoji{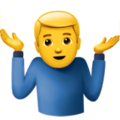} & \emoji{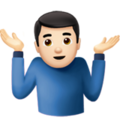} \\
\rowcolor[HTML]{EFEFEF} 
5 & My ancestors definitely gave me some guidance this year boy & Black & \emoji{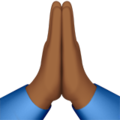} & \emoji{fold} & \emoji{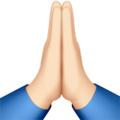} \\
6 & New month, New blessings, New beginnings & Black & \emoji{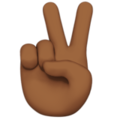} & \emoji{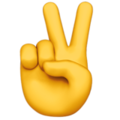} & \emoji{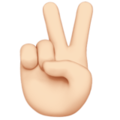} \\
\rowcolor[HTML]{EFEFEF} 
7 & \begin{tabular}[c]{@{}r@{}}Shout out to everyone that's fasting and had to use the\\ central line today\end{tabular} & Black & \emoji{v4} & \emoji{v} & \emoji{v1} \\
8 & Wish I had the money to dip this country & Black & \emoji{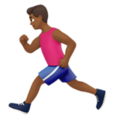} & \emoji{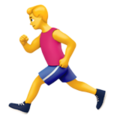} & \emoji{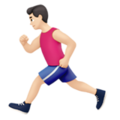} \\
\rowcolor[HTML]{EFEFEF} 
9 & Creed is a great film & Neutral & \emoji{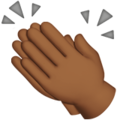} & \emoji{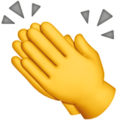} & \emoji{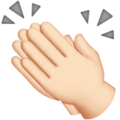} \\
10 & Feeling so brand new right now & Neutral & \emoji{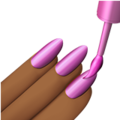} & \emoji{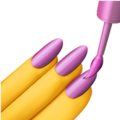} & \emoji{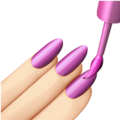} \\
\rowcolor[HTML]{EFEFEF} 
11 & Fool me once, shame on you. Fool me twice, shame on me & Neutral & \emoji{shrug4} & \emoji{shrug} & \emoji{shrug1} \\
12 & Got me feeling like i can get any polo & Neutral & \emoji{raising4} & \emoji{raising} & \emoji{raising1} \\
\rowcolor[HTML]{EFEFEF} 
13 & Looking out the window was my tune & Neutral & \emoji{raising4} & \emoji{raising} & \emoji{raising1} \\
14 & Next step manager then owner. By 28 just watch me & Neutral & \emoji{tip4} & \emoji{tip} & \emoji{tip1} \\
\rowcolor[HTML]{EFEFEF} 
15 & \begin{tabular}[c]{@{}r@{}}No matter the path you take, you're going to have to work hard. So\\ you might as well work hard for what you're really passionate about\end{tabular} & Neutral & \emoji{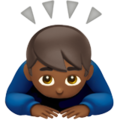} & \emoji{bow} & \emoji{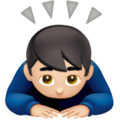} \\
16 & \begin{tabular}[c]{@{}r@{}}Stay positive. Stay focused. Stop worrying what could go wrong.\\ Instead, focus on what can go right. Be a warrior, not a worrier\end{tabular} & Neutral & \emoji{thumb4} & \emoji{thumb} & \emoji{thumb1} \\
\rowcolor[HTML]{EFEFEF} 
17 & \begin{tabular}[c]{@{}r@{}}Does blowing on tea actually make it cooler or is it just fun\\ to make the waves in your cup?\end{tabular} & White & \emoji{shrug4} & \emoji{shrug} & \emoji{shrug1} \\
18 & Fourth consecutive weekend of long haul flying. & White & \emoji{shrug4} & \emoji{shrug} & \emoji{shrug1} \\
\rowcolor[HTML]{EFEFEF} 
19 & I am ageing a year every minute & White & \emoji{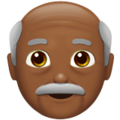} & \emoji{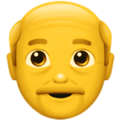} & \emoji{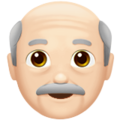} \\
20 & Need to get out! What's everyone doing tonight? & White & \emoji{bow4} & \emoji{bow} & \emoji{bow1} \\
\rowcolor[HTML]{EFEFEF} 
21 & Never too old for Christmas & White & \emoji{ok4} & \emoji{ok} & \emoji{ok1} \\
22 & Premiership games on a Friday the start of something very special & White & \emoji{raising4} & \emoji{raising} & \emoji{raising1} \\
\rowcolor[HTML]{EFEFEF} 
23 & The nap that I had was perfect & White & \emoji{ok4} & \emoji{ok} & \emoji{ok1} \\
24 & \begin{tabular}[c]{@{}r@{}}Today's spirit animal is the man on my train who has\\ commandeered a six seater area to lie down and go to sleep\end{tabular} & White & \emoji{clap4} & \emoji{clap} & \emoji{clap1} \\ \bottomrule
\end{tabular}%
}
\caption{The 24 text stimuli used in all experiments, with the three emoji used for each one. The text's linguistic bias, as determined through norming, is also shown.}
\label{appendix:stimuli}
\end{table}

\end{document}